\documentclass[runningheads]{llncs}
\usepackage[T1]{fontenc}
\usepackage{graphicx}
\usepackage{color}
\usepackage{cite}
\usepackage{subfigure}
\usepackage{booktabs}
\usepackage{amsmath}

\usepackage{multirow}
\usepackage{hyperref}
\usepackage{orcidlink}
\usepackage[utf8]{inputenc}
\usepackage[marginal]{footmisc}

\begin{document}
\title{Robust 3D Brain MRI Inpainting with Random Masking Augmentation\thanks{Corresponding author: Ying Weng.}}

%
%
\author{Juexin Zhang \orcidlink{0000-0001-9086-7342} \and
Ying Weng \orcidlink{0000-0003-4338-713X} \and
Ke Chen \orcidlink{0000-0002-2046-0034}}
\authorrunning{J. Zhang et al.}
%
\institute{University of Nottingham Ningbo China, Ningbo 315100, China\\
\email{\{juexin.zhang, ying.weng, ke.chen2\}@nottingham.edu.cn}\\}
\maketitle              
\begin{abstract}
The ASNR-MICCAI BraTS-Inpainting Challenge was established to mitigate dataset biases that limit deep learning models in the quantitative analysis of brain tumor MRI. This paper details our submission to the 2025 challenge, a novel deep learning framework for synthesizing healthy tissue in 3D scans. The core of our method is a U-Net architecture trained to inpaint synthetically corrupted regions, enhanced with a random masking augmentation strategy to improve generalization. Quantitative evaluation confirmed the efficacy of our approach, yielding an SSIM of 0.873$\pm$0.004, a PSNR of 24.996$\pm$4.694, and an MSE of 0.005$\pm$0.087 on the validation set. On the final online test set, our method achieved an SSIM of 0.919$\pm$0.088, a PSNR of 26.932$\pm$5.057, and an RMSE of 0.052$\pm$0.026. This performance secured first place in the BraTS-Inpainting 2025 challenge and surpassed the winning solutions from the 2023 and 2024 competitions on the official leaderboard.
\end{abstract}

\section{Introduction}

The quantitative analysis of brain tumors, particularly high-grade gliomas, from multi-modal Magnetic Resonance Imaging (MRI) is a cornerstone of modern neuro-oncology. It provides critical information for diagnosis, surgical planning, radiotherapy guidance, and monitoring treatment response. In recent years, deep learning models have achieved state-of-the-art performance on well-defined tasks such as tumor segmentation. However, the performance and generalizability of these models are intrinsically dependent on access to large, diverse, and accurately annotated datasets. This dependency reveals a fundamental challenge: the very pathology we aim to analyze introduces significant biases into the data and the standard computational workflows used to process it.

A fundamental challenge in neuro-oncological image analysis stems from a critical data void: for any given patient, a corresponding "healthy" scan from a pre-pathological state is almost never available. Patients typically undergo their first MRI only after the onset of symptoms, meaning their anatomical baseline is already compromised. This lack of a patient-specific, ground-truth healthy reference is the root cause of the "pathology bias" that confounds downstream algorithms. Without this reference, it is difficult to accurately quantify the true extent of anatomical deformation or to train models that can robustly differentiate pathological changes from normal inter-subject variability.

However, this challenge also illuminates a powerful new opportunity for generative modeling. If one could accurately synthesize the missing healthy anatomy for a patient, it would create a pristine digital canvas. This "anatomical canvas" would not only serve as an ideal reference for tasks like registration but would also enable a paradigm-shifting approach to data augmentation: generative pathology transplantation. This goes far beyond simple geometric transformations. By first inpainting the original tumor to create a healthy proxy, one can then programmatically synthesize and implant different tumors—of varying types, sizes, and growth patterns—onto the same patient's unique anatomical substrate.

\begin{figure}[t]
    \centering
    \includegraphics[width = \textwidth]{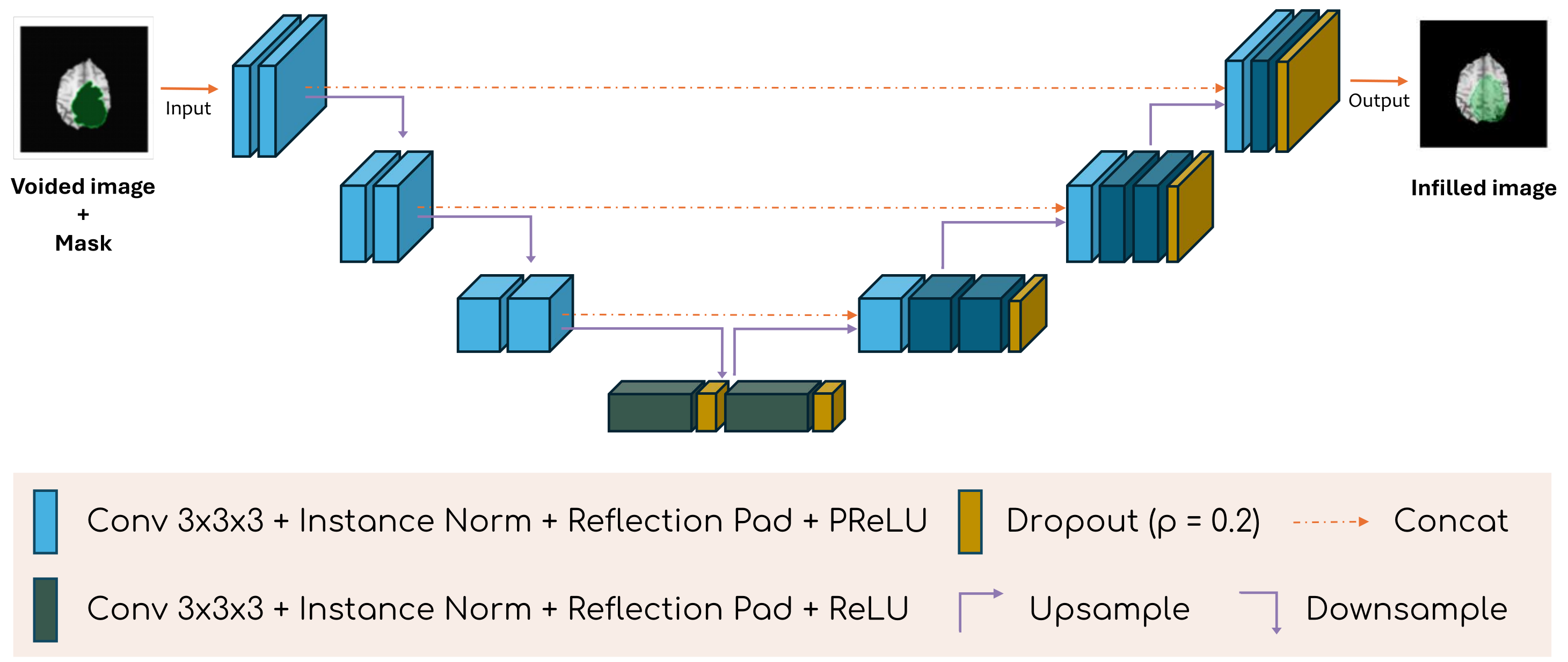}
    \caption{The figure illustrates the architecture of our U-Net model.}
    \label{unet}
\end{figure}

Consider a real-world clinical scenario: a dataset may be rich in cases of ring-enhancing glioblastoma but contain very few examples of diffuse, non-enhancing astrocytomas. A model trained on this biased data will naturally perform poorly on the rarer subtype. Using our proposed framework, we can take a scan of a patient with a glioblastoma, computationally "resect" the tumor via inpainting to generate a healthy version of their brain, and then synthetically "graft" a realistic, non-enhancing astrocytoma into the same location. This process generates a highly valuable, perfectly co-registered data pair: a specific patient's anatomy with two different, clinically relevant pathologies. By repeating this process, we can synthetically enrich datasets with rare disease manifestations, creating highly controlled counterfactuals that are essential for developing next-generation AI models that are not only accurate but also robustly generalizable across the entire spectrum of disease expression.

As our contribution to the ASNR-MICCAI BraTS Local Synthesis of Tissue via Inpainting (BraTS-Inpainting) Challenge, we propose a method to synthesize a subject-specific, "healthy" anatomical proxy from pathological MRI scans using a deep learning model with a U-Net backbone. The remainder of this paper is structured as follows: The BraTS dataset and the methodologies related to the U-Net like model are described in Section \ref{methods}. Section \ref{results} presents the experimental methods of the proposed model, and the paper is concluded in Section \ref{conclusion}.

\section{Methods}\label{methods}
\subsection{Dataset}

Our dataset extends the methodology of the BraTS-Local-Inpainting dataset \cite{kofler2023brain} to create a more comprehensive training resource. The dataset is built from 1251 T1-weighted MRI scans from the BraTS-GLI 2023 collection \cite{baid2021rsna}, each featuring tumor annotations approved by expert neuroradiologists.

A key distinction of our work is the generation of five unique healthy tissue masks for each source image. We first identified healthy brain tissue spatially separated from the tumor using the algorithm from \cite{kofler2023brain}. These masks were then augmented through random mirroring and rotation to create five distinct versions per scan, aiming to improve model generalization. All MRI volumes and masks were standardized to dimensions of 240×240×155. Each training sample consists of the following five components:

\begin{itemize}
\item t1n: The original ground truth T1-weighted image.
\item t1n-voided: The t1n image with regions corresponding to the healthy-mask and unhealthy-mask occluded.
\item healthy-mask: A binary mask identifying a unique, augmented region of healthy tissue.
\item unhealthy-mask: The binary mask of the expert-annotated tumor region.
\item mask: A combined binary mask representing the union of the healthy-mask and unhealthy-mask.
\end{itemize}

\subsection{Pre-processing}

We began with the BraTS 2021 GLI dataset, which had already undergone standard pre-processing: co-registration to a common anatomical template, resampling to a uniform $1mm^3$ isotropic resolution, and skull-stripping. We then applied our own processing pipeline. First, we normalized the images in two stages, scaling them to a $[0, 1]$ range by dividing by their maximum intensity value, and subsequently to a $[-1, 1]$ range. Following normalization, all MRI scans and masks were cropped to a size of $208\times208\times144$. Finally, to generate the output, our model's predictions on these cropped patches are stitched together with the original T1-weighted MRI.

\begin{figure}[t]
    \centering
    \subfigure[BraTS-GLI-00114-000 (best): SSIM 0.999069, PSNR 38.939751, MSE 0.000128]{
    \includegraphics[width=\textwidth]{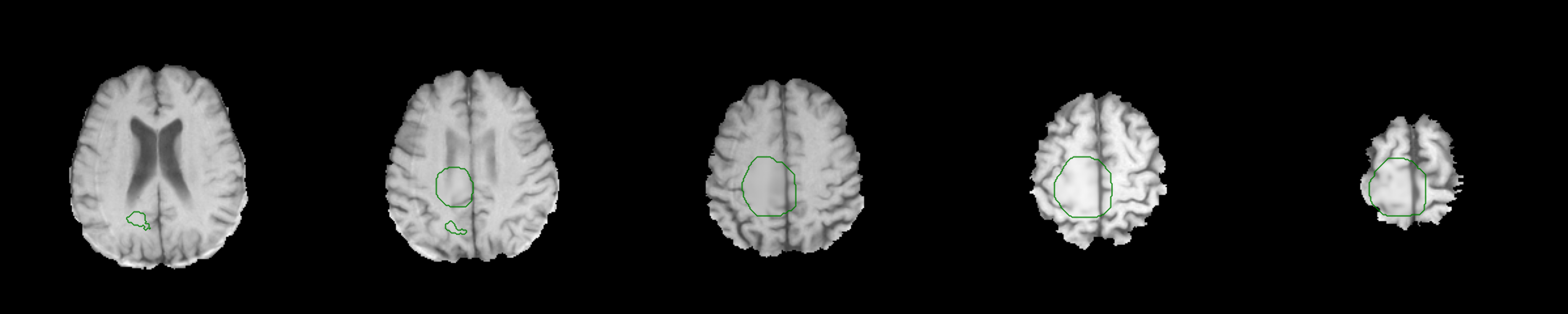}\label{sub1}}
    \subfigure[BraTS-GLI-01773-000 (median): SSIM 0.906654, PSNR 22.563231, MSE 0.005542]{
    \includegraphics[width=\textwidth]{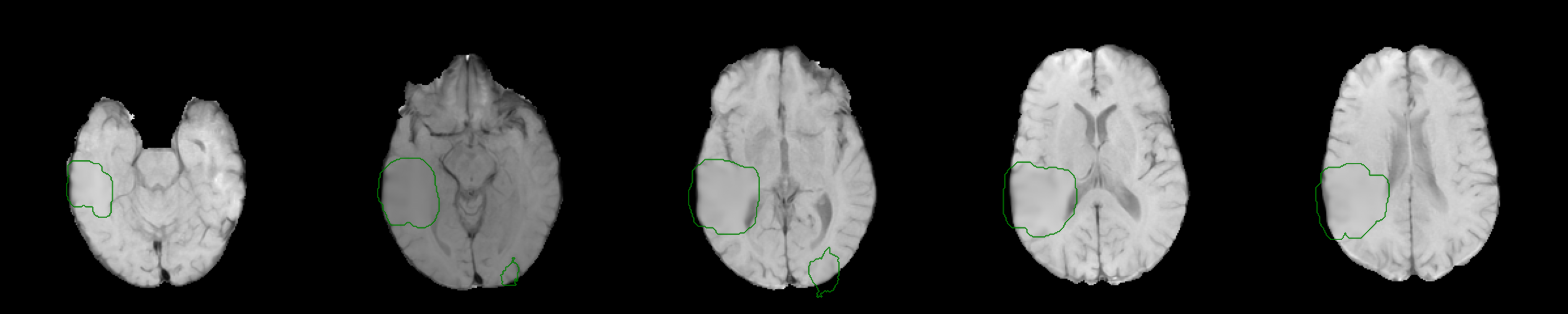}\label{sub2}}
    \subfigure[BraTS-GLI-00467-000 (worst): SSIM 0.709323, PSNR 16.336607, MSE 0.023246]{
    \includegraphics[width=\textwidth]{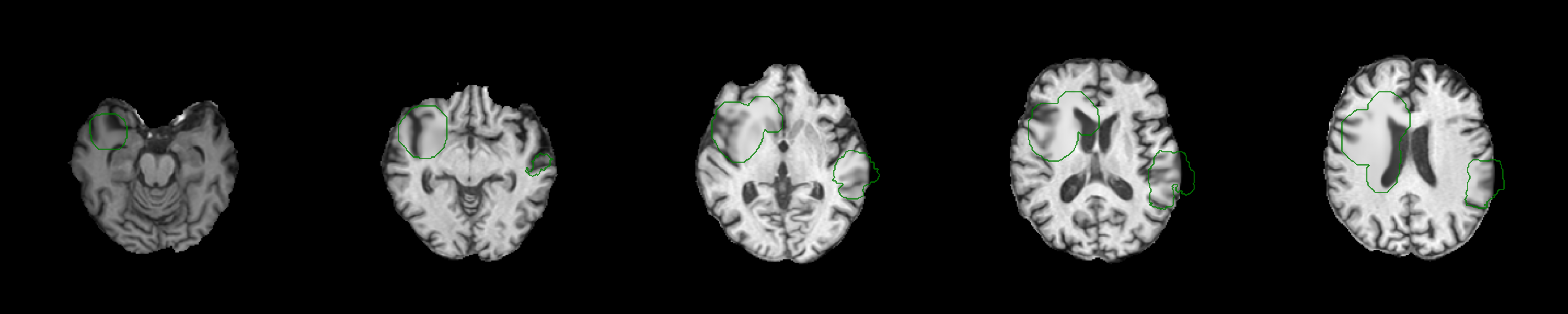}\label{sub3}}

    \caption{Qualitative results of our model's infilling performance on validation MRI scans, showcasing the best (Fig \ref{sub1}), median (Fig \ref{sub2}), and worst (Fig \ref{sub3}) cases. The green masks indicate the inpainted regions, which contained both healthy and unhealthy tissues as these were not explicitly labeled.}
    \label{fig1}
\end{figure}

\subsection{Data Augmentation}

To address the common issue of overfitting in high-capacity deep learning models and improve generalization to unseen data, we employ a robust data augmentation strategy. As mentioned above, we generate five unique healthy tissue masks using the algorithm described in \cite{kofler2023brain} for each MRI scan. These masks are created by applying a set of random transformations, including mirroring and rotation. Specifically, mirroring is applied independently to each dimension with a probability of 50\%, while random rotations between 0 ° and 360 ° are performed on both the XY and YZ planes. Although some overlap between the five masks for a given scan is expected, the resulting variability in mask shape, location, and size is crucial for training a more robust and generalized model.

\subsection{Network Architecture}

For the task of synthesizing healthy tissue, we propose a model based on the U-Net architecture \cite{ronneberger2015u}. As depicted in Figure~\ref{unet}, the network utilizes an encoder-decoder structure comprising three downsampling blocks, a central bridge block, and three upsampling blocks.

Each block contains two 3D convolutional layers with a $3 \times 3 \times 3$ kernel. We employ Parametric ReLU (PReLU) as the activation function in the downsampling and upsampling blocks, while the standard ReLU is used in the bridge. Instance normalization is applied after each convolutional layer to stabilize training. The number of feature channels starts at 32, doubling with each downsampling step and halving with each upsampling step. Skip connections are integrated to pass features from the encoder stages to their corresponding decoder stages, preserving low-level details. To mitigate overfitting, dropout with a rate of 0.2 is applied in the bridge and upsampling blocks. The model accepts a t1n-voided image and its associated mask as input, and it outputs an infilled image.

\subsection{Loss Function}

The model is trained using a composite loss function, which is a weighted sum of the Mean Absolute Error (MAE) and the Structural Similarity Index Measure (SSIM) \cite{wang2004image}. This hybrid design was motivated by our observation that SSIM alone performs poorly in preserving masked regions and can introduce artifacts at mask boundaries. To address this, the MAE component is calculated exclusively on the healthy regions of the ground truth ($GT$) and the generated image ($I$), while the SSIM component is computed on the entire images to maintain overall structural coherence.

The loss functions are formulated as follows:
\begin{flalign}
    & \quad MAE(x, y) = \frac{1}{m}\sum_{i=1}^{m}|y_{i}-f(x_{i})| && \\
    & \quad SSIM(x, y) = \frac{(2\mu_{x}\mu_{y}+c_{1})(2\sigma_{xy}+c_{2})}{(\mu_{x}^{2}+\mu_{y}^{2}+c_{1})(\sigma_{x}^{2}+\sigma_{y}^{2}+c_{2})} && \\
    & \quad Loss(I, GT) = \lambda_{1} \cdot MAE(I, GT) + \lambda_{2} \cdot SSIM(I, GT) &&
\end{flalign}

\begin{figure}[t]
    \centering
    \subfigure[BraTS-GLI-00467-000 (2023 winner)]{
    \includegraphics[width=\textwidth]{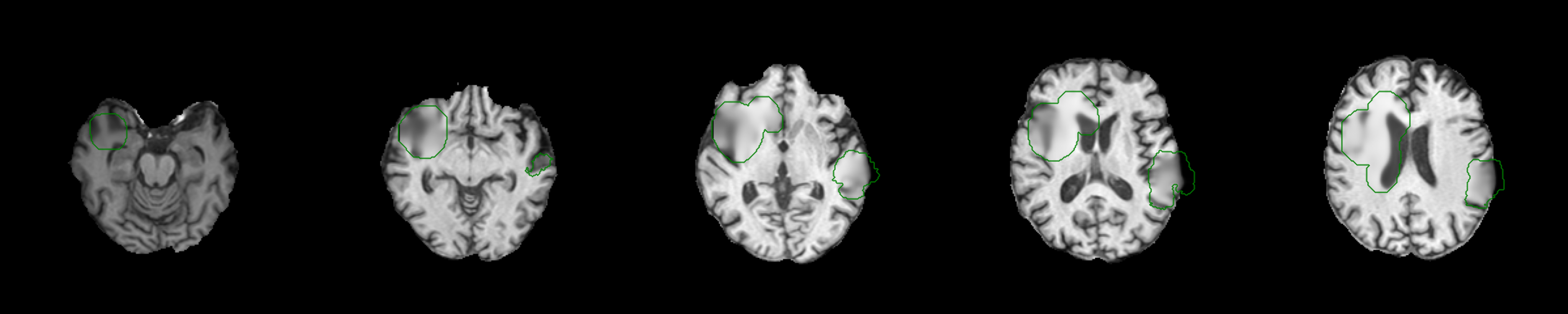}}
    \subfigure[BraTS-GLI-00467-000 (2024 winner)]{
    \includegraphics[width=\textwidth]{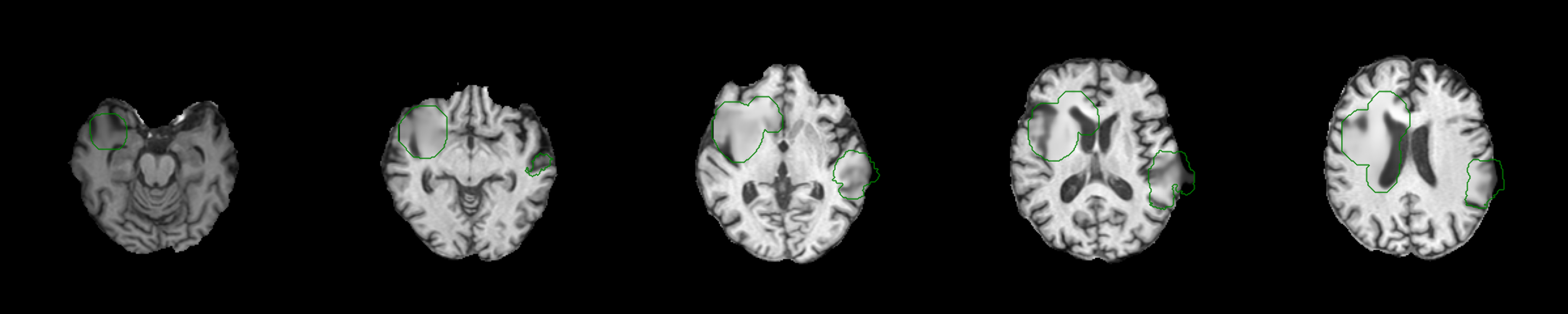}}
    \subfigure[BraTS-GLI-00467-000 (ours)]{
    \includegraphics[width=\textwidth]{imgs/BraTS-GLI-00467-000.png}}

    \caption{Qualitative comparison of our method's worst-performing case against the winning methods of the BraTS 2023 and 2024 challenges.}
    \label{fig2}
\end{figure}

\section{Experiment Results}\label{results}

\subsection{Evaluation Metrics}
The performance of our model is quantitatively assessed by comparing the generated healthy regions to the ground truth data. We use three standard metrics for this evaluation: Structural Similarity Index Measure (SSIM), Peak-Signal-to-Noise-Ratio (PSNR), and Mean-Square-Error (MSE). Notably, this evaluation is performed only on the healthy regions defined by the ground truth masks.

\subsection{Experiment Settings}
We utilized 5-fold cross-validation to select hyperparameters. Models were trained for a maximum of 500 epochs, and for each fold, we saved the checkpoint corresponding to the lowest validation loss. We employed the Adam optimizer with an initial learning rate of $1 \times 10^{-4}$ and betas of $(0.9, 0.999)$. The loss weights $\lambda_{1}$ and $\lambda_{2}$ were both set to 1. During the validation phase, the healthy mask region of the ground truth image is normalized to a $[0, 1]$ range based on the maximum intensity value across both healthy and unhealthy regions.

\subsection{Validation Phase}

Table \ref{tab1} summarizes the quantitative performance of our method on the BraTS-Local-Inpainting validation dataset, benchmarked against the winning solutions of the 2023 \cite{10.1007/978-3-031-76163-8_21} and 2024 challenges \cite{zhang2025u}. All metrics were computed using the official Sage Bionetworks Synapse online evaluation platform. To complement this analysis, Figure \ref{fig1} provides qualitative results, illustrating the best, median, and worst inpainting cases from the validation set. The evaluation protocol has two key constraints: 1) the ground truth is withheld, precluding direct visual comparison, and 2) scoring is confined to healthy tissue, although inpainting masks cover both healthy and pathological regions.

Visually, our model effectively captures fine-grained textures and generates plausible tissue structures that integrate well with the surrounding context. However, the inpainted regions exhibit some blurriness, an artifact most prominent in low-intensity areas (Fig. \ref{sub2}, fourth column; Fig. \ref{sub3}, first column). We attribute this to the Mean Absolute Error (MAE) loss, which can incentivize the model to predict the mean of plausible solutions, leading to oversmoothed textures. We also illustrate the worst-performing case of our method against the winner solution of 2023 and 2024 in Figure \ref{fig2}.

\begin{table}[ht]
\centering
\caption{Online validation data results.}
\label{tab1}
\begin{tabular}{c @{\hspace{1cm}} l @{\hspace{0.5cm}} c c c c c}
\toprule
\multicolumn{1}{l}{\multirow{2}{*}{}} & \multirow{2}{*}{} & \multirow{2}{*}{MSE} & \multirow{2}{*}{PSNR} & \multirow{2}{*}{SSIM} \\
\multicolumn{1}{l}{} &  &  &  &  \\
 \midrule
\multirow{10}{*}{2023 winner \cite{10.1007/978-3-031-76163-8_21}} & \multirow{2}{*}{Mean} & \multirow{2}{*}{0.00931688} & \multirow{2}{*}{21.4458628} & \multirow{2}{*}{0.8119463} \\
 &  &  &  &  \\
 & \multirow{2}{*}{Standard deviation} & \multirow{2}{*}{0.00645289} & \multirow{2}{*}{3.44400102} & \multirow{2}{*}{0.11350122} \\
 &  &  &  &  \\
 & \multirow{2}{*}{25 quantile} & \multirow{2}{*}{0.00489901} & \multirow{2}{*}{18.4753408} & \multirow{2}{*}{0.69751903} \\
 &  &  &  &  \\
 & \multirow{2}{*}{Median} & \multirow{2}{*}{0.00815745} & \multirow{2}{*}{20.8844547} & \multirow{2}{*}{0.82023758} \\
 &  &  &  &  \\
 & \multirow{2}{*}{75 quantile} & \multirow{2}{*}{0.01239707} & \multirow{2}{*}{23.689291} & \multirow{2}{*}{0.89072207} \\
 &  &  &  &  \\
  \midrule
\multirow{10}{*}{2024 winner \cite{zhang2025u}} & \multirow{2}{*}{Mean} & \multirow{2}{*}{0.00650362} & \multirow{2}{*}{23.3814246} & \multirow{2}{*}{0.84116632} \\
 &  &  &  &  \\
 & \multirow{2}{*}{Standard deviation} & \multirow{2}{*}{0.00466064} & \multirow{2}{*}{4.26449611} & \multirow{2}{*}{0.10317845} \\
 &  &  &  &  \\
 & \multirow{2}{*}{25 quantile} & \multirow{2}{*}{0.00278909} & \multirow{2}{*}{20.3874092} & \multirow{2}{*}{0.75842395} \\
 &  &  &  &  \\
 & \multirow{2}{*}{Median} & \multirow{2}{*}{0.00579681} & \multirow{2}{*}{22.3681049} & \multirow{2}{*}{0.84412175} \\
 &  &  &  &  \\
 & \multirow{2}{*}{75 quantile} & \multirow{2}{*}{0.00914667} & \multirow{2}{*}{25.5454702} & \multirow{2}{*}{0.92018622} \\
 &  &  &  &  \\
 \midrule
\multirow{10}{*}{Ours} & \multirow{2}{*}{Mean} & \multirow{2}{*}{\textbf{0.00476023}} & \multirow{2}{*}{\textbf{24.9959218}} & \multirow{2}{*}{\textbf{0.87300897}} \\
 &  &  &  &  \\
 & \multirow{2}{*}{Standard deviation} & \multirow{2}{*}{0.00360885} & \multirow{2}{*}{4.69427685} & \multirow{2}{*}{0.08699671} \\
 &  &  &  &  \\
 & \multirow{2}{*}{25quantile} & \multirow{2}{*}{0.00188717} & \multirow{2}{*}{21.7267790} & \multirow{2}{*}{0.80683365} \\
 &  &  &  &  \\
 & \multirow{2}{*}{Median} & \multirow{2}{*}{0.00405384} & \multirow{2}{*}{23.9213314} & \multirow{2}{*}{0.87929922} \\
 &  &  &  &  \\
 & \multirow{2}{*}{75 quantile} & \multirow{2}{*}{0.00671933} & \multirow{2}{*}{27.2419672} & \multirow{2}{*}{0.94228190} \\
 &  &  &  &  \\

\bottomrule
\end{tabular}
\end{table}
\clearpage
\subsection{Test Phase}
We present our model's overall performance on the test set in Table \ref{tab2}. Due to restricted access, the organizers executed the inference runs, which prevents us from providing visualizations of the infilled images. Despite this lack of visual evidence, the quantitative results confirm our model's robust performance and strong generalization capabilities.

\begin{table}[ht]
\centering
\caption{Performance Metrics on the Test Set.}
\label{tab2}
\begin{tabular}{lcccc}
\toprule
 & & SSIM & PSNR & RMSE \\
\midrule
Mean & & 0.91928125 & 26.9321548 & 0.05162604\\
Standard deviation & & 0.08843877 & 5.0565417 & 0.02610840\\
\bottomrule
\end{tabular}
\end{table}

\section{Conclusion\label{conclusion}}

In this paper, we presented a novel deep learning framework for synthesizing healthy brain tissue in pathological MRI scans, our winning submission to the ASNR-MICCAI BraTS-Inpainting 2025 Challenge. Our method, centered on a U-Net architecture enhanced with a random masking augmentation strategy, demonstrated state-of-the-art performance. The quantitative results, achieving an SSIM of 0.919, a PSNR of 26.932, and an RMSE of 0.052 on the final test set, not only secured first place but also surpassed the winning entries from previous years. While our qualitative analysis identified minor blurriness as an area for improvement, the overall success underscores the model's robustness and its potential to mitigate pathology-induced data bias. This work represents a significant step forward in generating high-fidelity anatomical proxies, opening new possibilities for data augmentation and the development of more generalizable AI models in neuro-oncological image analysis.

\section*{Acknowledgements}
This work was supported by Ningbo Major Science \& Technology Project under Grant 2022Z126. (Corresponding: Ying Weng.)

%
%
%
\bibliographystyle{splncs04}
\bibliography{ref.bib}

\end{document}